%% file: acl.tex
\pdfoutput=1

\documentclass[11pt]{article}

\usepackage{acl}

\usepackage{times}
\usepackage{latexsym}

\usepackage[T1]{fontenc}

\usepackage[utf8]{inputenc}

\usepackage{microtype}

\usepackage[export]{adjustbox}

\usepackage{graphicx}
\usepackage{multirow}
\usepackage{adjustbox}
\usepackage{tabularx}

\usepackage{booktabs}
\usepackage{amsmath}
\usepackage{bbm}
\usepackage{caption}
\usepackage{subcaption}

\DeclareMathOperator*{\argmax}{argmax}

\title{
    Fantastically Ordered Prompts and Where to Find Them: \\
    Overcoming Few-Shot Prompt Order Sensitivity
}

\author{Yao Lu$^{\dagger}$ \quad Max Bartolo$^{\dagger}$ \quad Alastair Moore$^{\ddagger}$ \quad Sebastian Riedel$^{\dagger}$ \quad Pontus Stenetorp$^{\dagger}$\\
    $^{\dagger}$University College London \quad
    $^{\ddagger}$Mishcon de Reya LLP\\
    \texttt{\{yao.lu,m.bartolo,s.riedel,p.stenetorp\}@cs.ucl.ac.uk} \\
    \texttt{alastair.moore@mishcon.com}
    }

\date{}

\begin{document}
\maketitle
\input{main}

\bibliography{acl}
\bibliographystyle{acl_natbib}

\clearpage

\appendix

\input{appendix}

\end{document}

%% file: main.tex
\begin{abstract}

When primed with only a handful of training samples, very large, pretrained language models such as GPT-3 have shown competitive results when compared to fully-supervised, fine-tuned, large, pretrained language models.
We demonstrate that the order in which the samples are provided can make the difference between near state-of-the-art and random guess performance: essentially some permutations are ``fantastic'' and some not.
We analyse this phenomenon in detail, establishing that: it is present across model sizes (even for the largest current models), it is not related to a specific subset of samples, and that a given good permutation for one model is not transferable to another.
While one could use a development set to determine which permutations are performant, this would deviate from the true few-shot setting as it requires additional annotated data.
Instead, we use the generative nature of language models to construct an artificial development set and based on entropy statistics of the candidate permutations on this set, we identify performant prompts.
Our method yields a 13\% relative improvement for GPT-family models across eleven different established text classification tasks.

\end{abstract}

\section{Introduction}

Large pretrained language models~\cite[PLMs,][]{devlin2019bert,peters2018deep,raffel2020exploring,liu2019roberta,yang2019xlnet,Radford2019LanguageMA} have shown remarkable performance when conditioned with an appropriate textual context~\cite{petroni-etal-2019-language,petroni2020context,jiang2020can, shin2020autoprompt,davison2019commonsense}. For example, when conditioned on a long document and a ``TL;DR:'' token, they can generate a summary of said document, and when provided a partial question (``The theory of relativity was developed by \_\_''), they can generate the correct answer.
Perhaps most strikingly, when primed with a context consisting of very few training examples, they produce text classification results that can match those of fully supervised models.
This type of few shot setting, is commonly referred to as ``In-context Learning''~\cite{brown2020language}.

\begin{figure}
    \centering
    \includegraphics[width=0.8\columnwidth]{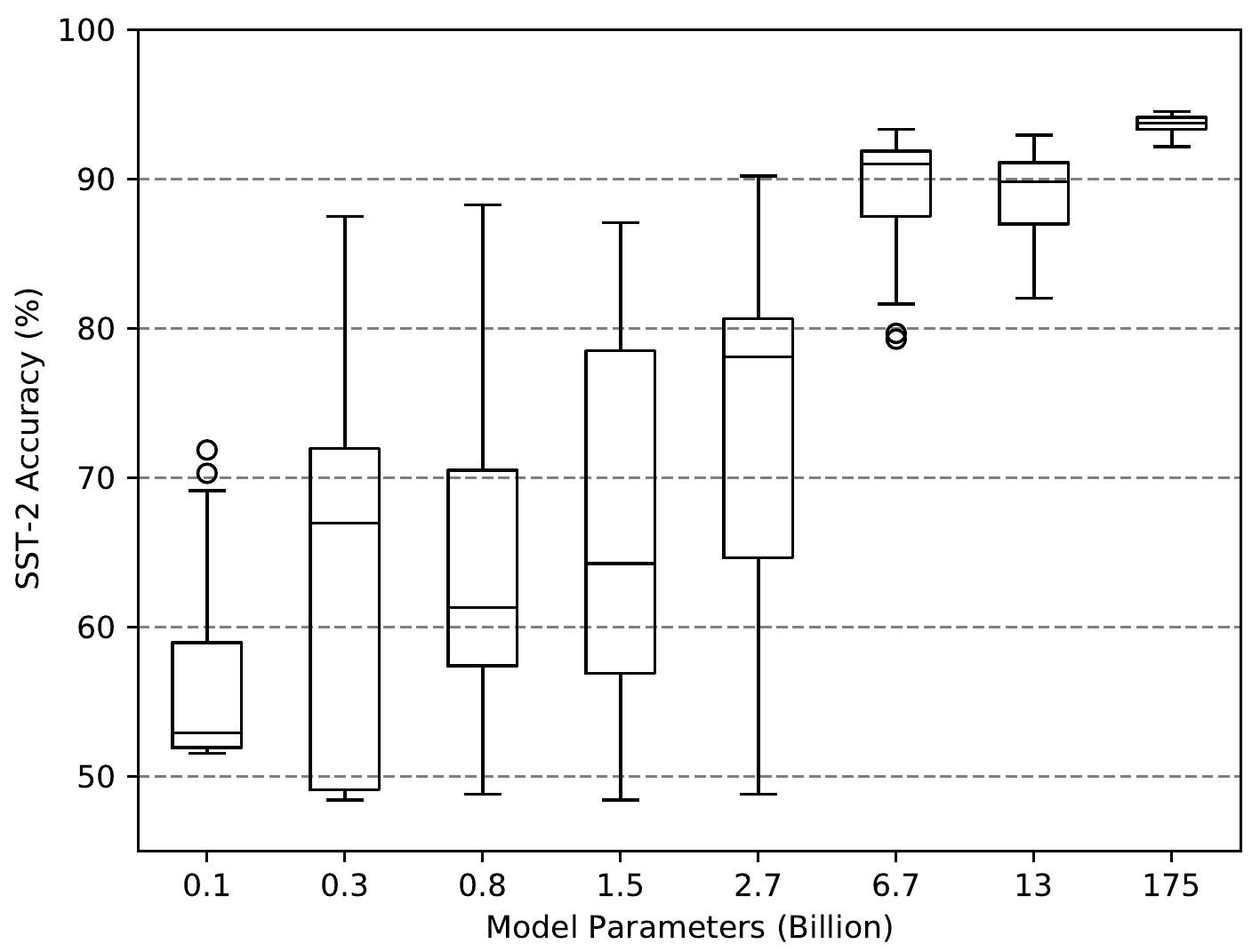}
    \includegraphics[width=0.8\columnwidth]{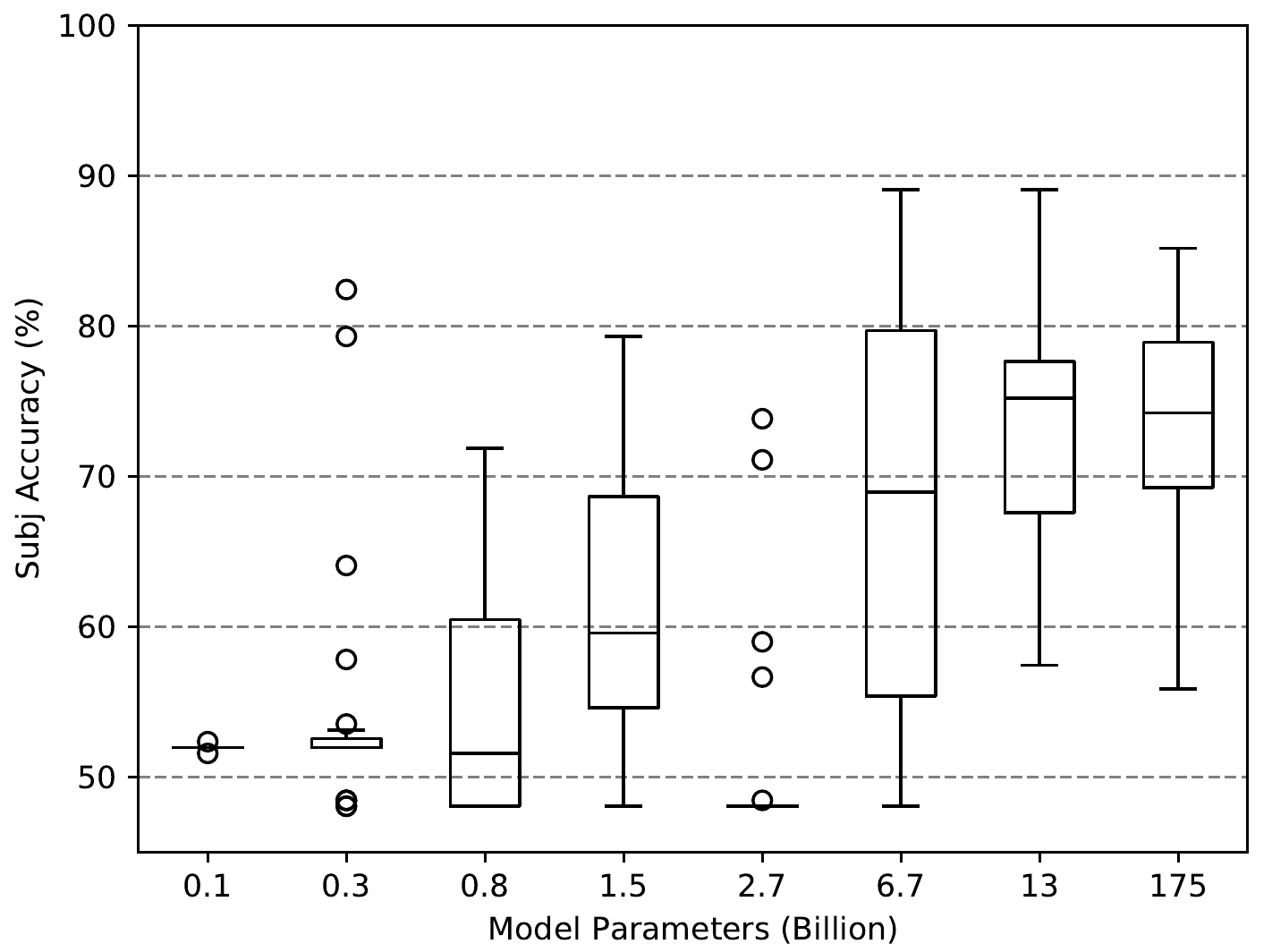}

    \caption{
        Four-shot performance for 24 different sample orders across different sizes of GPT-family models (GPT-2 and GPT-3) for the SST-2 and Subj datasets.
    }
    \label{fig:sst2-permutation}
\end{figure}

A core component of in-context learning is the text-based prompt that serves as the context.
Composing a prompt requires: (i) text linearisation using a template; and (ii) training sample concatenation (See Table~\ref{tab:prompt_construction_example} for an example).
It has been established that the structure of the template has a large impact on performance~\cite{shin2020autoprompt,gao2020making,schick2020pet,jiang2020can}.
However, to the best of our knowledge, no work has studied the effect of the sample ordering on In-context Learning performance.

Perhaps counter-intuitively, we find that the right sample order can make as much of a difference as the right template. 
\input{tables/example_of_prompt}
As can be seen in Figure~\ref{fig:sst2-permutation}, some permutations have comparable performance (over 85\% accuracy) to supervised training for sentiment classification, while others perform close to random (around 50\%).
This order sensitivity is universal across models, and although increasing the model size somewhat addresses it, the problem is still present for some text classification tasks (Subj in Figure~\ref{fig:sst2-permutation}) for models with billions of parameters.

In our analysis, we find no common denominator between performant sample orders and that they are not transferable across different model sizes and tasks.
In a fully-supervised setting, we could rely on a development set to select among sample orders.
However, this is not desirable in a few-shot setting where the size of the development set is very limited, even unavailable~\cite{perez2021true} .
Instead, we use the generative nature of language models to construct an unlabelled artificial development set and refer to it as a \textit{probing set}. %
As the probing set is unlabelled, we use the predicted label distribution statistics and propose entropy-based metrics to measure the quality of candidate prompts.%
Experimental results show that we can achieve on average 13\% relative improvement across eleven different established text classification tasks across all different sizes (four orders of magnitude) of PLMs.

To summarise, our contributions are as follows:
\begin{enumerate}
    \item{We study order sensitivity for In-context Learning, which we show is crucial for the success of pretrained language models for few-shot learning.}
    \item{We propose a simple, generation-based probing method to identify performant prompts without requiring additional data.}
    \item{Our probing method is universally applicable and effective across different sizes of pretrained language models and for different types of datasets -- achieving on average a 13\% relative improvement over a wide range of tasks.}
\end{enumerate}

\section{Order Sensitivity and Prompt Design}\label{sec:order}
\begin{figure}
    \centering
    \includegraphics[width=\columnwidth]{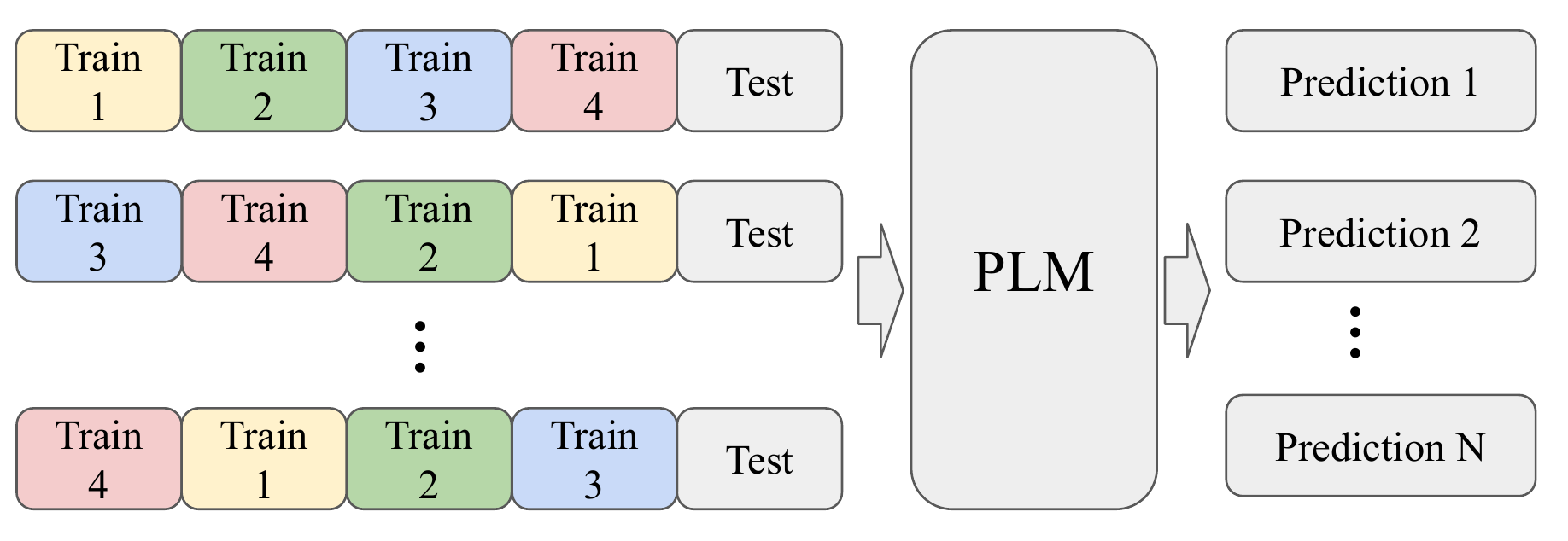}
    \caption{
        Training sample permutations for the In-context Learning setting.
        The concatenation of training samples as well as test data transforms the classification task into a sequence generation task.
    }
    \label{fig:permutation_pipeline}
\end{figure}

In this section, we study the relationship between permutation performance and various factors. 
For the ease of visualisation, we use a fixed random subset of four samples with a balanced label distribution from the SST-2 dataset and consider all 24 possible sample order permutations.
This setup is illustrated in Figure~\ref{fig:permutation_pipeline}. 
We also test five randomly-selected sets of examples and summarised variance statistics in the experiment section (Section~\ref{sec:main-result}).

\paragraph{Although beneficial, increasing model size does not guarantee low variance} We evaluate the order permutations for four different sizes of GPT-2 (0.1B--1.5B)%
\footnote{We can also refer these models as GPT2-base, GPT2-medium, GPT2-Large, and GPT2-XL.} 
and GPT-3 (2.7B--175B).
As we can observe in Figure~\ref{fig:sst2-permutation}, models can obtain remarkable few-shot performance.
We see that the GPT2-XL (1.5B) model can even surpass 90\% accuracy given just four samples.
This result is comparable to those of supervised models trained on more than 60,000 samples.
However, the performance variation of different permutations remain a big issue, especially for ``smaller'' models.%
\footnote{
    The smallest model in our experiment is the same size as BERT-base.
}
The same model can exhibit nearly perfect behaviour given one sample order, but then fall back to be on par with a random baseline for another.
While increasing the model size (by a few order of magnitudes) can sometimes alleviate the issue, it still cannot resolve it entirely (especially if we consider tasks other than SST-2).
In contrast, different initialisations of supervised fine-tuning approaches typically result in less than 1\% standard deviation for their test set performance~\cite{gao2020making}.

\begin{figure}
    \centering
    \includegraphics[width=\columnwidth]{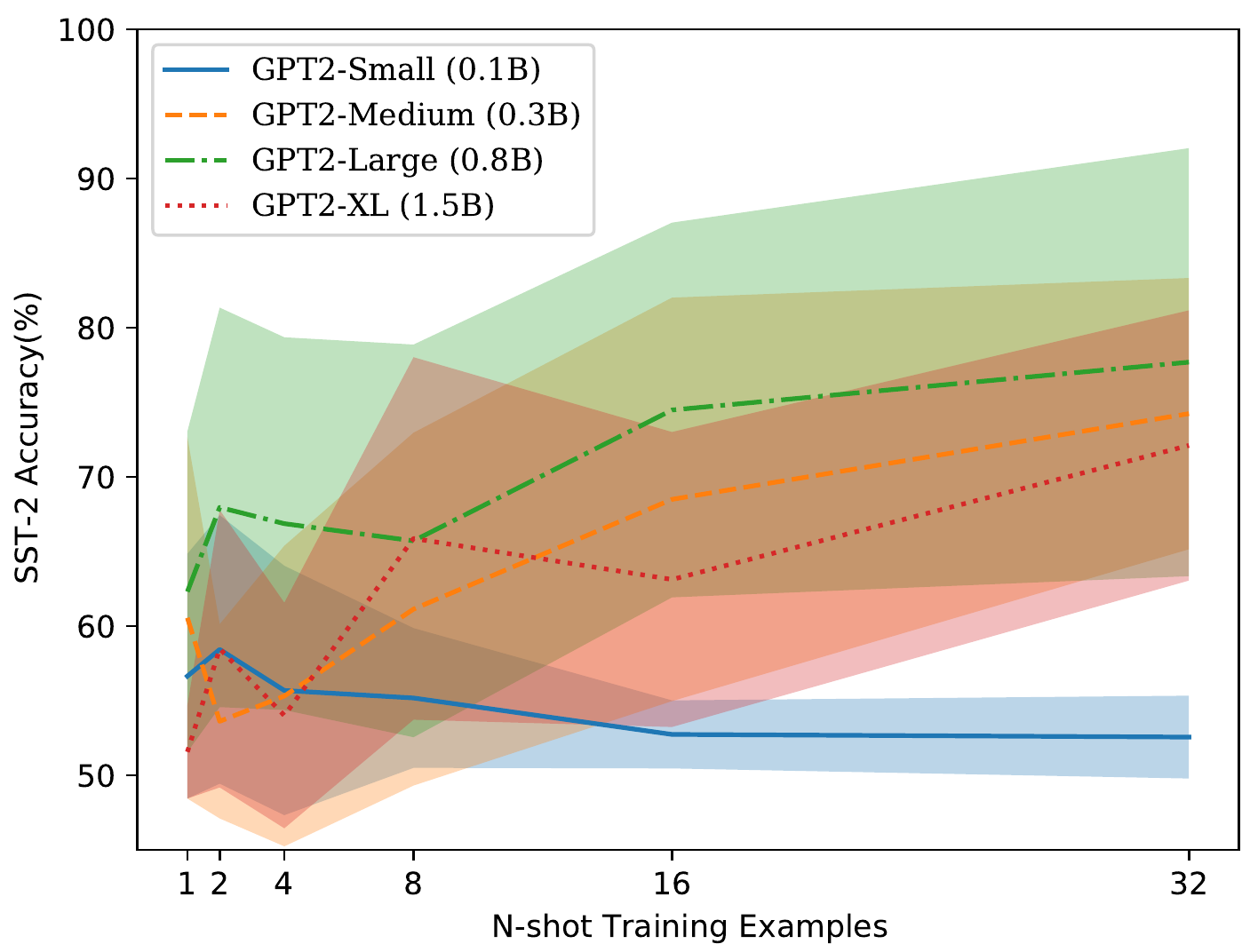}
    \caption{Order sensitivity using different numbers of training samples.}
    \label{fig:moreshot}
\end{figure}

\paragraph{Adding training samples does not significantly reduce variance}
To further explore the order sensitivity of few-shot prompts, we increase the number of training samples and then sample a subset of at most 24 different orderings.%
\footnote{
    Bounded at the lower limit by the total number of samples given, and at the upper limit as there can be up to $64!$ possible orders.
}
We use the GPT2 family models for this experiment.
In Figure~\ref{fig:moreshot}, we can observe that increasing the number of training samples leads to increases in performance.
However, a high level of variance remains, even with a large number of samples and can even increase.
Based on this, we draw the conclusion that order sensitivity is likely to be a fundamental issue of In-context Learning regardless of the number of training samples.

\begin{figure}
    \centering
    \includegraphics[width=\columnwidth]{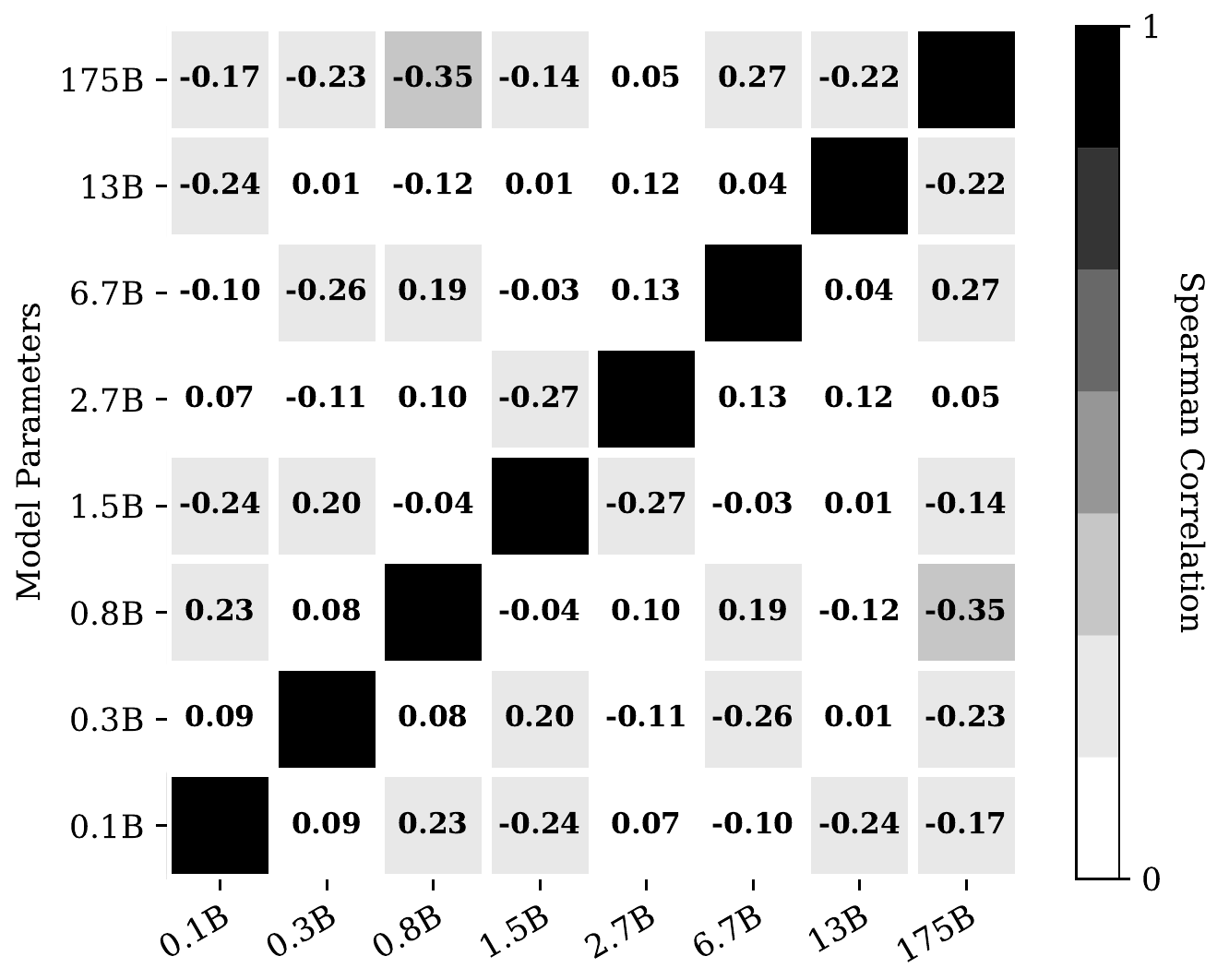}
    \caption{
         Training sample permutation performance correlation across different models.
    }
    \label{fig:correlation}
\end{figure}

\paragraph{Performant prompts are not transferable across models}
We find that a specific permutation's performance may drop from 88.7\% to 51.6\% by changing the underlying model from GPT2-XL (1.5B) to GPT2-Large (0.8B).
This suggests that a particular permutation working well for one model does not imply that it will provide good results for another model. 
To validate this hypothesis, we use all possible order permutations of the four samples as prompts -- 24 in total.
We then perform prediction conditioned on each of these prompts for different models and calculate the pairwise Spearman's rank correlation coefficient between the scores.
These results are shown in Figure~\ref{fig:correlation}.

If there is a common pattern for performant prompts, we should then be able to observe high correlation across models.
However, the behaviour of permutations is seemingly random even across different sizes of the same model.
For example, the 175B and 2.7B model only has a correlation of 0.05, this means a good permutation for the 2.7B model is in no way guaranteed that it will also yield good performance for the 175B model.
\begin{figure}
    \centering
    \includegraphics[width=\columnwidth]{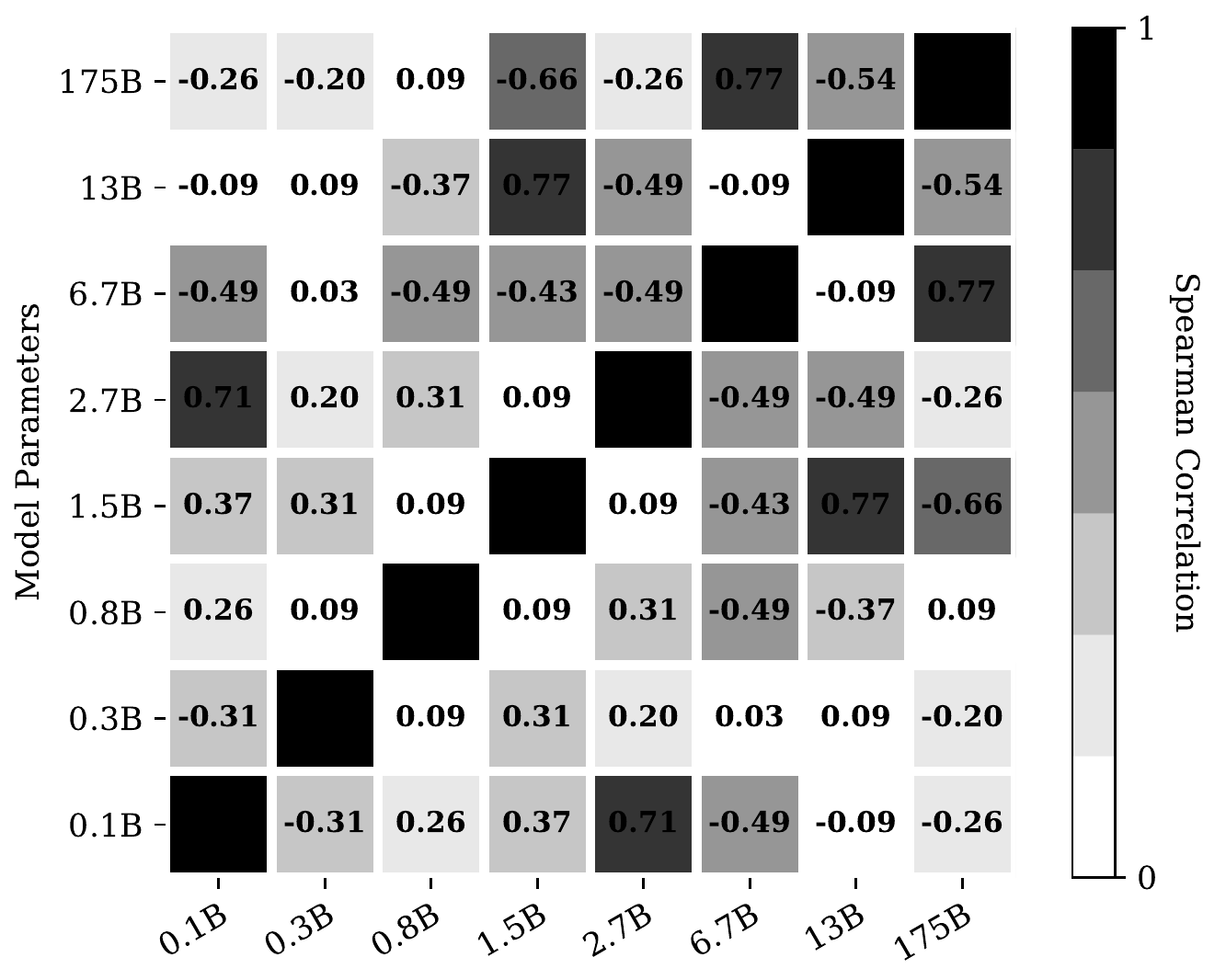}
    \caption{Training label pattern permutation performance correlation across different models.}
    \label{fig:correaltion_pattern}
\end{figure}

\paragraph{{Performant label orderings are not consistent across models}} In addition to training example ordering, we also explore label ordering for training prompts. We use all patterns of the above-mentioned full permutations -- six different label patterns.%
\footnote{NNPP, NPNP, NPPN, PNNP, PNPN, PPNN, where P/N respectively denotes positive/negative} 
We then compute the pairwise Spearman correlation across different models as described in the previous paragraph. 
As shown in Figure~\ref{fig:correaltion_pattern}, the behaviour of label orderings is once again seemingly random across different sizes of the same model.
It is thus not possible to identify a label ordering that is performant across different models.

\begin{figure}

    \centering
    \includegraphics[width=3.8cm,valign=t]{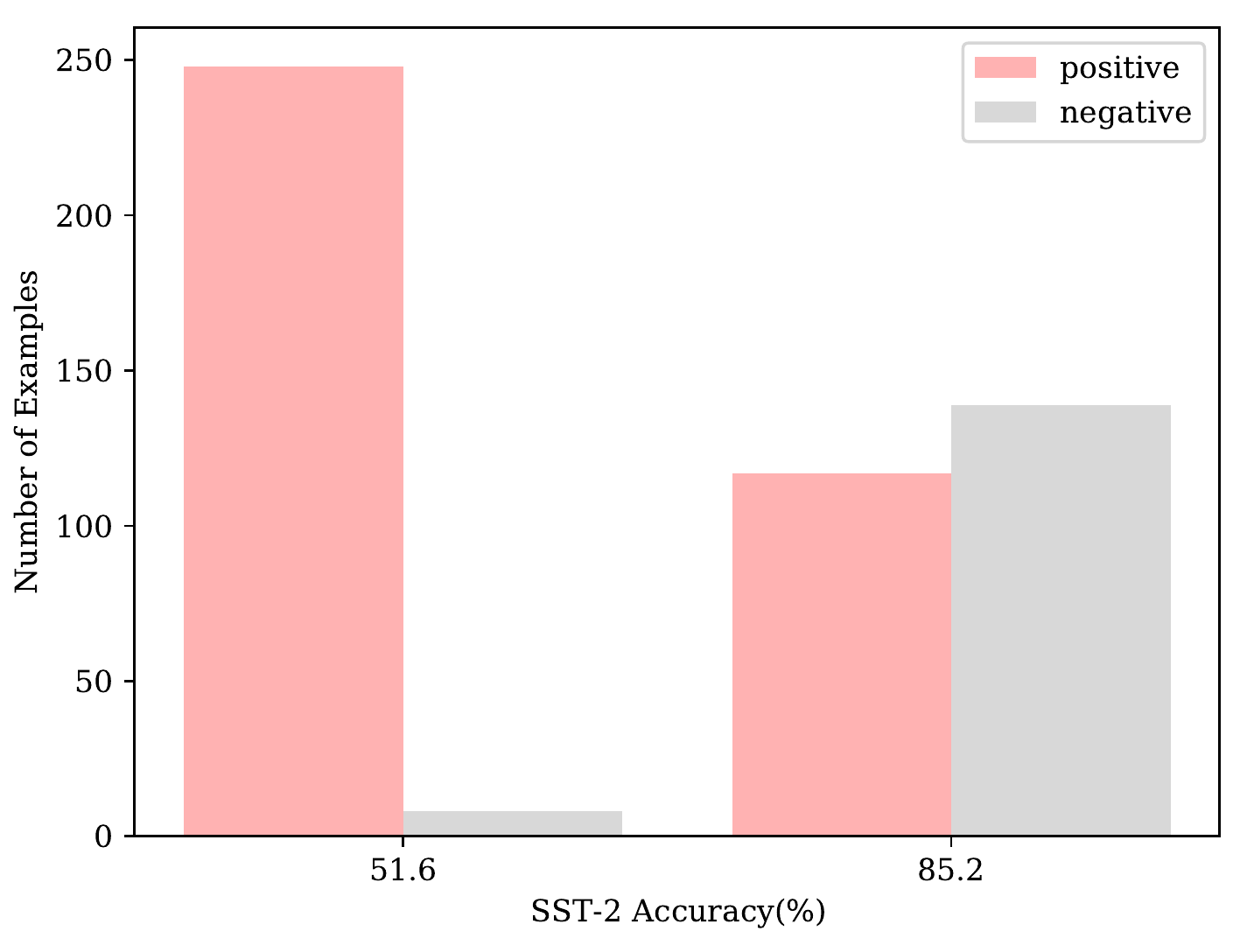}
    \includegraphics[width=3.8cm,valign=t]{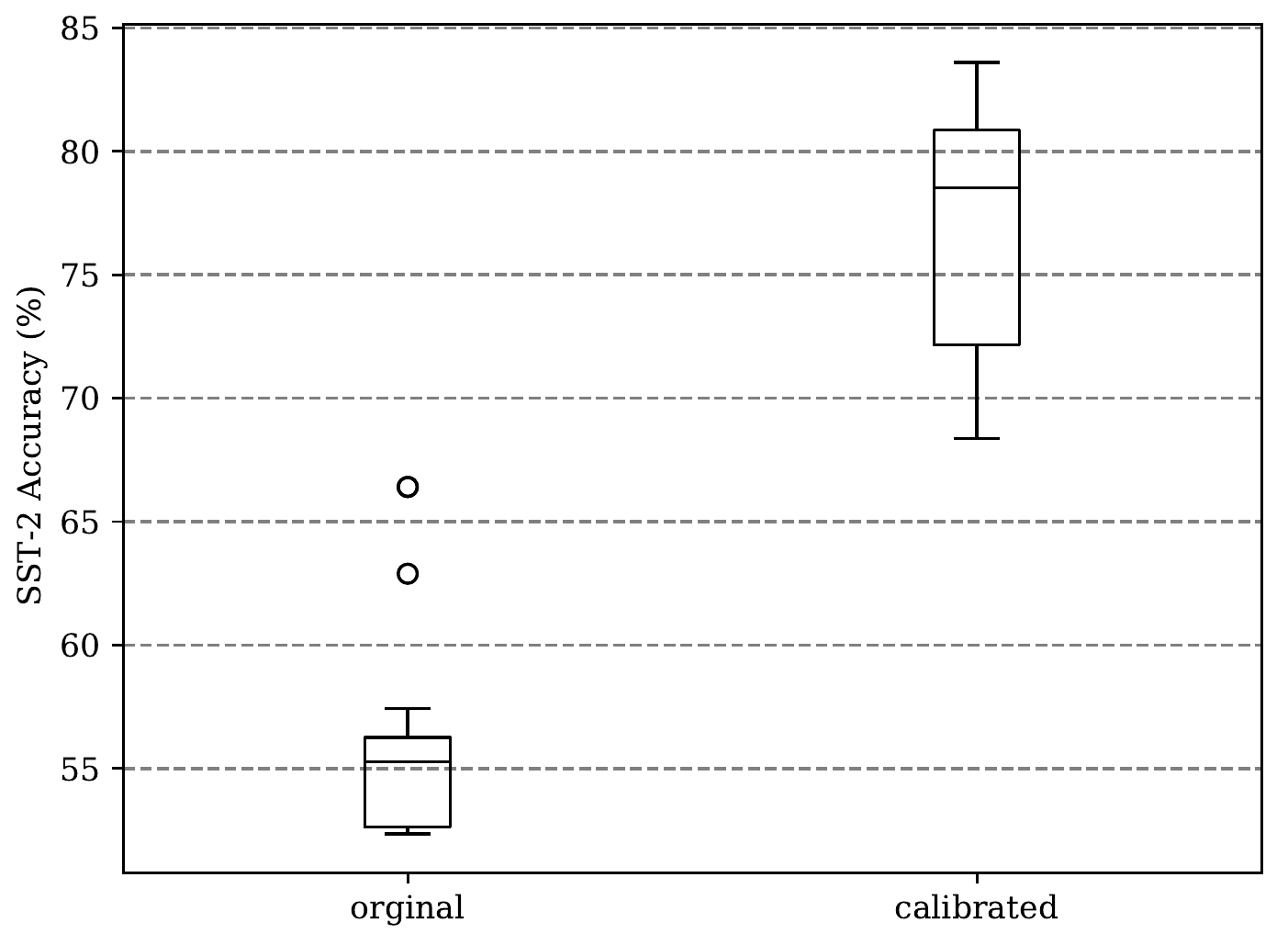}
    \caption{
        Left: Predicted SST-2 label distribution under different prompts.
        Right: 2-shot calibrated performance~\cite{zhao2021calibrate} of all possible permutations on GPT2-XL (1.5B).
    }
    \label{fig:label_dist}
\end{figure}

\paragraph{Degenerate behaviour of bad prompts}
We perform error analysis across performant and non-performant prompts and observe that the majority of failing prompts suffer from highly unbalanced predicted label distributions~(Figure~\ref{fig:label_dist}, left).
An intuitive way to address this would be by calibrating the output distribution, along the lines of \citet{zhao2021calibrate}.
However, we find that although calibration leads to much higher performance, the variance remains high~(Figure~\ref{fig:label_dist}, right).

\begin{figure*}
    \centering
    \includegraphics[width=2.08\columnwidth]{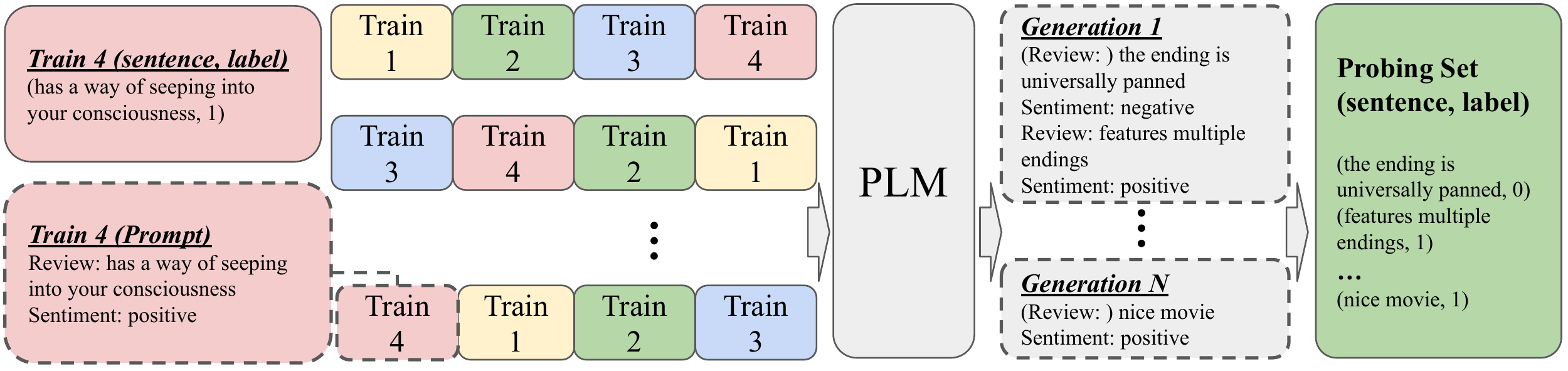}
    \caption{
        Our probing set construction method, showing the various possible ordering permutations of the randomly selected training samples, the resulting generation for each permutation, and the concatenation of each into a probing set.
        Note that we discard the generated labels, as there is no guarantee that these generated labels are correct.
    }
    \label{fig:generation_probing}
\end{figure*}

\section{Methodology}

The previous section demonstrates that prompt order can have a substantial effect on performance, with some orderings of the same prompts for the same model providing random performance, and other ``better'' orderings providing performance competitive with supervised approaches.
This suggests that there could be various ways of selecting prompt orders to achieve better performance, but the challenge is to do so automatically and without the need for additional labels (e.g., a development set).

Hence, in this section, we explore the question of: ``How can we automatically generate a `probing set' to find performant prompt orderings''?
We approach this by: (i) for a randomly-selected set of training samples, we use every possible ordering permutation of this set as candidates; (ii) constructing a \textit{probing set} by querying the language model using all candidate prompts as context; and (iii) use this probing set to identify the best ordering by ranking them using a probing metric.
\subsection{Sampling from the Language Model to Construct a Probing Set}

We propose a simple methodology to automatically construct a ``probing set'', by directly sampling from the language model itself. This approach makes it possible to generate probing sets automatically, without access to any additional data.
Concretely, given a set of training samples $S=\{(x_i, y_i)\}, i=1,\cdots,n$, where $x_i$ and $y_i$ denote the sentence and label of the $i^\text{th}$ training sample.
We then define a transformation $\mathcal{T}$, mapping each sample into natural language space, such that $t_i = \mathcal{T}(x_i, y_i)$. $t_i$ is therefore a text sequence of the $i^\text{th}$ training sample using the template defined by $\mathcal{T}$.
In this work, we use a simple transformation function $\mathcal{T}$ such that $\mathcal{T}(x_i, y_i) = \text{input:}x_i \text{ type:} y_i$. 
This transforms each sample into a standard format sentence, which linearises each element in the set into natural language space defined as $S^{'}=\{t_i\}, i=1,\cdots,n$.

We then define a full permutation function group of $n$ training samples, $\mathcal{F} = \{f_m\}, m=1,\cdots, n!$, where each function $f_m$ takes $S^{'}$ as input and outputs $c_m$: the concatenation of a unique permutation.
In our case, sampling four training samples at random gives up to 24 possible ordering permutations of the transformed samples.

For each prompt candidate $c_m$, we then sample from the language model to obtain the probing sequence $g_m \sim P(\cdot| c_m; \theta) $, where $\theta$ denotes the parameters of the pretrained language model.
We stop decoding from the language model upon generating the special end-of-sentence token defined by a template, or reach the generation length limit.
Our probing set construction method is illustrated in  Figure~\ref{fig:generation_probing}, where the objective is to generate a probing set that shares a similar distribution to the training samples.

We run this sampling process for all possible prompt ordering permutations and extract probing samples from them ($\mathcal{T}^{-1}(g)$).
Then gather extracted samples together to form the probing set $D = \mathcal{T}^{-1}(g_1) \oplus ... \oplus \mathcal{T}^{-1}(g_{n!})$.
Although the probing set contains predicted label for each sentence, there is no guarantee on the validity of these labels.
Therefore, we discard them from the probing set as we are only interested in sampling probes from the language model corresponding to the input distribution.

\subsection{Probing Metrics}

Once we have constructed a probing set for a given set of samples, we can now use that probing set to identify the best possible prompt ordering for that particular sample set.
Here, we explore two methods for selecting the best ordering: Global Entropy~(GlobalE), and Local Entropy~(LocalE).

\paragraph{Global Entropy (GlobalE)}{
The motivation behind GlobalE is to identify prompts of specific sample orderings that avoid the issue of extremely unbalanced predictions (as we have previously established it as key problem for non-performant prompts).
We compute the predicted label $\hat{y}_i$ for data point $(x^{'}_i, y^{'}_i)$ under context $c_m$ as follows:%
\begin{equation}
    \hat{y}_{i,m} = \argmax_{v \in V} P(v|c_m\oplus \mathcal{T}(x^{'}_i);\theta )
\end{equation}

For each label $v \in V$ (where $V$ denotes the target label set), we compute the label probability over the probing set as:
\begin{equation}
p^{v}_{m} = \frac{\sum_{i} \mathbbm{1}_{\{\hat{y}_{i,m} = v\}}}{|D|}
\end{equation}

We then use the predicted category label entropy as the GlobalE score for $c_m$ as follows:
\begin{equation}
    \text{GlobalE}_{m} = \sum_{v \in V} -p^{v}_{m} \log{p^{v}_{m}}
\end{equation}
}

\paragraph{Local Entropy (LocalE)} {
The motivation behind LocalE is that if a model is overly confident for all probing inputs, then it is likely that the model is not behaving as desired.
At the very least, it is poorly calibrated, which could also be an indication of a poor capability to appropriately differentiate between classes.
Similar to the GlobalE computation, we calculate the prediction probability of a data point $(x^{'}_i, y^{'}_i)$ over the target labels $v \in V$ under context $c_m$, as follows:
\begin{equation}
    p^{v}_{i,m} =P_{(x^{'}_i, y^{'}_i) \sim D}(v|c_m\oplus \mathcal{T}(x^{'}_i);\theta ), v \in V
\end{equation}

We then calculate the average prediction entropy per data point as the LocalE score:

\begin{equation}
    \text{LocalE}_{m} = \frac{ \sum_{i} \sum_{v \in V} - p^{v}_{i,m} \log p^{v}_{i,m} }{|D|}
\end{equation}
}
As we now have a way to score each prompt ordering, based on its effect against the probing set, we can rank each prompt ordering by performance as measured by GlobalE or LocalE respectively.

\section{Experimental Setup}

\input{tables/main_result}

We use four different sizes of GPT-2~\citep{Radford2019LanguageMA} (with 0.1B, 0.3B, 0.8B, and 1.5B parameteers) and two sizes of GPT-3~\citep{brown2020language} (with 2.7B, and 175B parameters).
Due to limited context window size (up to 1024 word-pieces for the GPT-2 series of models), we use a 4-shot setting for all datasets except AGNews and DBPedia.
Our experiments are based on the open-source checkpoints of GPT-2 models and access to the OpenAI GPT-3 API.\footnote{https://openai.com/api/}
For probing set generation, we restrict the maximum generation length to 128.
We also use sampling with a temperature, $t$, of 2, and we also make use of block $n$-gram repetitions~\cite{paulus2018deep} to encourage diverse generation.

We use 24 different permutations for each set of randomly selected training samples and use 5 different sets (except for GPT-3 with 175B parameters, where we only do two sets with 12 different permutation due to the high monetary cost) for each experiment, giving a total of 120 runs.
We report the mean and standard deviation of the corresponding evaluation metric over 5 different sets.%

For performant prompt selection, we rank candidate prompts using the LocalE and GlobalE probing metrics over the automatically generated probing set.
We then select top $k$ samples ranked by highest entropy values, where $k=4$ in our experiments, of the available 24 permutations as performant prompts.
Finally, we use these performant prompts to evaluate performance on various datasets and demonstrate both better performance and reduced variance.
We also provide results for a majority baseline, which always predicts the majority label in the dataset, as a lower-bound of performance.
We also provide an oracle to show the upper-bound of performance by selecting the top four performant orderings based on prompt performance on the validation set.

\subsection{Evaluation Datasets}
Similar to previous work~\cite{gao2020making, zhao2021calibrate}, we use eleven text classification datasets ranging from sentiment classification to textual entailment.
Further details of the datasets are provided in the Appendix.
For evaluation, we sub-sample 256 samples of the validation sets for all datasets to control for the GPT-3 inference costs as it requires the usage of a monetary paid-for API.

\section{Results}\label{sec:main-result}
We report experimental results in Table~\ref{tab:main_result} and observe consistent improvements for both LocalE and GlobalE across all tasks. 

\paragraph{Entropy-based probing is effective for performant prompt selection regardless of model size}
{
We find that GlobalE achieves, on average, a 13\% relative improvement across the eleven different sentence classification tasks in comparison to prompts that do not make use of probing.
LocalE provides results slightly inferior to GlobalE, with an average 9.6\% relative improvement over the baseline model.
Our selected performant prompts also demonstrate considerably lower variance than using all candidate prompts.
}

\begin{figure}[h]
    \centering
\includegraphics[width=1.02\columnwidth]{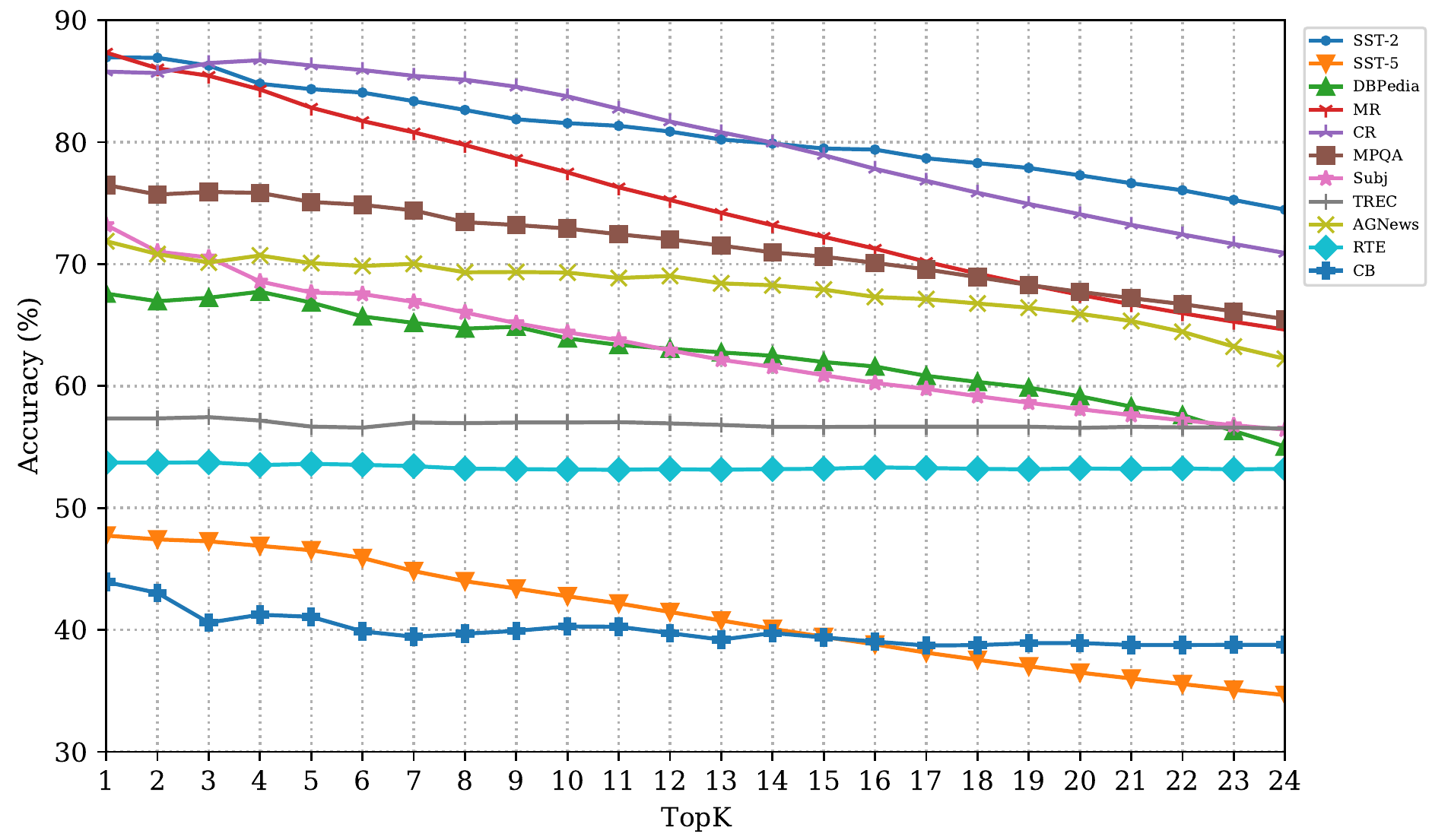}

    \caption{Average performance of different Top $K$ permutation selection on GPT2-Large (0.8B)}
    \label{fig:topk}
\end{figure}

\paragraph{Ranking using Entropy-based probing is robust}{
In Figure~\ref{fig:topk}, we visualise the average performance when varying $K$ for the top $K$ prompt selection.
$K=24$ corresponds to using all sampled prompt orders, which is equivalent to the baseline model performance in Table~\ref{tab:main_result}.
We can observe that the slope of curves are negative for all datasets, suggesting that our method can rank performant prompts effectively.
Though $K=1$ can provide good performance for most cases, in our experiments, we use $K=4$ as preliminary experiments indicated that it yielded stable performance across datasets.
}

\paragraph{Entropy-based probing is effective across templates}{
We evaluate Entropy-based probing for four different templates similar to~\citet{gao2020making} and~\citet{zhao2021calibrate} (Table~\ref{tab:different_template}) for the SST-2 dataset.
Experimental results in Table~\ref{tab:different-template-result} indicate that Entropy-based probing is valid for different templates.
We also observe that the randomness across different templates is similar to Section~\ref{sec:order}.
These findings suggest that Entropy-based probing is not sensitive to specific templates, as it consistently provides improvements for all cases.
}

\input{tables/change_template}
\input{tables/different_template}

\paragraph{Performant permutation selection is a safe option for In-context Learning}
{
We find that for models that suffer from high prompt variance, our prompt selection process can show large improvements -- up to 30\% relative improvement.
Furthermore, for tasks with low initial prompt performance variance, our method does not negatively impact performance.
Our prompt selection provides marginal improvement at worse and on average a 13\% relative improvement in the most cases.
}

\paragraph{Sentence-pair tasks remain challenging for smaller-sized models even with performant permutation selection}

For the CB and RTE datasets, the performance of GPT-2 models is not significantly different from that of a random baseline.
Despite this, we find that our method for identifying performant prompts can still provide minimal performance gains, although these are still within the levels of a random guess or majority vote.
One reason for this could be that, for these particular sizes of models on these tasks, no good prompt exists.
As such, optimising the prompt is not particularly effective in this setting.
This is further supported by the observation that prompt selection can considerably improve performance on both CB and RTE at larger model sizes (particularly so for the GPT-3 175B parameter model). 
In fact, we find that prompt selection using GlobalE improves performance by 4.9\% for GPT-3 175B on CB.
This indicates that our method is widely applicable to all model sizes, and across all tasks, as long as they already possess some existing classification ability that can be improved through prompt design.

\paragraph{Entropy-based probing outperforms using subsets of the training data for tuning}
If one was not to rely on generation, an alternative approach to prompt selection could be to split the (limited) training data to form a validation set.
To compare against this approach, we split the 4-shot training samples (same setting as in Table~\ref{tab:main_result}) in half.
We then select the top four performing prompts using validation set performance.
As can be seen in Table~\ref{tab:cross_validation}, this approach consistently outperforms the baseline.
However, both Entropy-based probing methods consistently provides better performance across all model sizes.

\input{tables/cross_validation}

\section{Related Work}

\paragraph{Unified Interface Design for NLP} 
Most previous work focuses on shared-parameters models, pretrain on some tasks, then fine-tune for different tasks, e.g. ELMo~\cite{peters2018deep}, BERT~\cite{devlin2019bert}, etc. Eventually, leading to multiple task-specific models.
There has for some time been attempts to design a unified interface for NLP tasks~\cite{kumar2016ask, raffel2020exploring}.%
In parallel with these works, GPT-2~\cite{Radford2019LanguageMA} shows that appending trigger tokens (e.g. ``TL;DR'') at the end of language model input can cause language models to behave like summarisation models. The zero-shot capability of language models shows the potential to unify NLP tasks into a language modelling framework where fine-tuning is not necessary to achieve good performance. Furthermore, GPT-3~\cite{brown2020language} shows that task-agnostic, few-shot performance can be improved by scaling up language models. It can sometimes even become competitive with prior state-of-the-art fine-tuning approaches.%

\paragraph{Prompt Design for PLMs} The core challenge of prompt design is to convert training data (if it exists) into a text sequence. Most work on prompt design focuses on how to make prompts more compatible with language models. \citet{petroni-etal-2019-language} uses human effort to design natural language sentences and then perform token prediction given the input context. However, hand-crafted templates require significant human effort and is likely to end up with sub-optimal performance. Recent work has explored automatic template construction: ~\citet{schick2020pet} uses cloze-style tasks to construct templates,~\citet{gao2020making} uses an external language model to generate templates, and~\citet{shin2020autoprompt} uses gradient-guided search to find templates that maximise performance. \citet{jiang2020can} uses a mining-based method to create multiple diverse templates automatically. 

\paragraph{Order Sensitivity of Prompt Design}
\citet{gao2020making} demonstrated that finetuning-based approaches are not as order sensitive as In-context Learning.
Making use of a standard-size training set, \citet{liu2021makes} used nearest neighbour search to retrieve the most relevant training samples for a specific test sample.
They were successful in retrieving relevant samples and concluded that after retrieving them the order in which they are provided in the prompt has little to no effect on performance.
While our study is fundamentally different from theirs in that we do not make use of a standard-size training set, we do come to the opposite conclusion.
All previous work on prompt design focuses on the textual quality of the prompt and, to the best of our knowledge, none has studied order sensitivity in detail.
\paragraph{True Few-shot Learning}
\citet{perez2021true} evaluated few-shot capability of LMs when a held-out validation set is not available. Experimental result suggested that previous work overestimate the few-shot ability of LMs in this~(true few-shot learning) setting. Our work instead use the generative nature of language models to construct a probing set without relying on held-out examples. We show that our probing method is better than relying on held out examples (Figure~\ref{tab:cross_validation}) and thus enables true few-shot learning.

\section{Conclusion}
We have shown that few-shot prompts suffer from order sensitivity, in that for the same prompt the order in which samples are provided can make the difference between state-of-the-art and random performance.
In our analysis of the problem, we established that it is present across tasks, model sizes, prompt templates, samples, and number of training samples.
To alleviate this problem, we introduced a novel probing method that exploits the generative nature of language models to construct an artificial development set.
We were able to identity performant permutations using entropy-based statistics over this set, leading to an on average 13\% improvement across eleven text classification tasks.%

%% file: tables/example_of_prompt.tex
\begin{table}
\begin{center}
\centering
\resizebox{\columnwidth}{!}{%
\begin{tabular}{ll}
\toprule
 &  Example\\
 \midrule
training set & \begin{tabular}[c]{@{}l@{}}(the greatest musicians, 1)\\(redundant concept, 0)\end{tabular} \\
\midrule
linearization & \begin{tabular}[c]{@{}l@{}}{Review: the greatest musicians. Sentiment: positive}\\{Review: redundant concept. Sentiment: negative}\end{tabular}\\
\midrule
concatenation & \begin{tabular}[c]{@{}l@{}}Review: the greatest musicians. Sentiment: positive. Review:  redundant \\concept. Sentiment: negative\\ \textit{OR}\\Review: redundant concept. Sentiment: negative. Review: the greatest \\musicians. Sentiment: positive \end{tabular}\\

\bottomrule
\end{tabular}
}
\end{center}
\caption{Procedures for prompt construction.} 
\label{tab:prompt_construction_example}
\end{table}

%% file: tables/main_result.tex
\begin{table*}[!t]
\begin{center}
\centering
\resizebox{2.08\columnwidth}{!}{%
\begin{tabular}{lccccccccccc}
\toprule
      & SST-2 & SST-5 & DBPedia & MR & CR & MPQA & Subj & TREC & AGNews  & RTE & CB\\
\midrule
Majority  &  50.9 &  23.1 &  9.4   & 50.0    &  50.0   &  50.0   &  50.0   & 18.8  &  25.0   &  52.7   & 51.8 \\

Finetuning (Full) & 95.0 & 58.7 &  99.3 & 90.8 & 89.4 &  87.8 & 97.0 & 97.4 & 94.7 & 80.9 & 90.5 \\

\midrule
GPT-2 0.1B  &  $58.9_{7.8}$ &  $29.0_{4.9}$ & $44.9_{9.7}$   & $58.6_{7.6}$    &  $58.4_{6.4}$   &  $68.9_{7.1}$   &  $52.1_{0.7}$   & \textbf{49.2}$_{4.7}$  &  $50.8_{11.9}$   &  $49.7_{2.7}$   &  $50.1_{1.0}$\\
~  LocalE &  \textbf{65.2}$_{3.9}$ & $34.4_{3.4}$   & $53.3_{4.9}$ &  $66.0_{6.3}$ & \textbf{65.0}$_{3.4}$ & $72.5_{6.0}$    &  $52.9_{1.3}$   & $48.0_{3.9}$  & $61.0_{5.9}$   &  \textbf{53.0}$_{3.3}$   & $49.9_{1.6}$\\
~  GlobalE     &  $63.8_{5.8}$ & \textbf{35.8}$_{2.0}$  &  \textbf{56.1}$_{4.3}$ &  \textbf{66.4}$_{5.8} $  & $64.8_{2.7}$    & \textbf{73.5}$_{4.5}$    &  \textbf{53.0}$_{1.3}$  & $46.1_{3.7}$ &  \textbf{62.1}$_{5.7}$   &  \textbf{53.0}$_{3.0}$    & \textbf{50.3}$_{1.6}$ \\
~ Oracle & {73.5}$_{1.7}$ & {38.2}$_{4.0}$ & {60.5}$_{4.2}$ & {74.3}$_{4.9}$ & {70.8}$_{4.4}$ & {81.3}$_{2.5}$ & {55.2}$_{1.7}$ & {58.1}$_{4.3}$ & {70.3}$_{2.8}$ & {56.8}$_{2.0}$ & {52.1}$_{1.3}$\\

\midrule

GPT-2 0.3B &  $61.0_{13.2}$ & $25.9_{5.9}$  &   $51.7_{7.0}$  &  $54.2_{7.8}$   & $56.7_{9.4}$    & $54.5_{8.8}$  &  $54.4_{7.9}$   &  $52.6_{4.9}$  & $47.7_{10.6}$ &  $48.8_{2.6}$   & $50.2_{5.3}$ \\
~ LocalE &  $75.3_{4.6}$ & $31.0_{3.4}$  & $47.1_{3.7}$  & $65.2_{6.6}$  &  \textbf{70.9}$_{6.3}$   &  $67.6_{7.2}$   &  \textbf{66.7}$_{9.3}$   & $53.0_{3.9}$  & $51.2_{7.3}$     &  \textbf{51.8}$_{1.0}$   &  $47.1_{4.2}$\\

~ GlobalE     &  \textbf{78.7}$_{5.2}$  & \textbf{31.7}$_{5.2}$ &   \textbf{58.3}$_{5.4}$  & \textbf{67.0}$_{5.9}$ &  $70.7_{6.7}$   &  \textbf{68.3}$_{6.9}$  &  $65.8_{10.1}$   &  \textbf{53.3}$_{4.6}$ & \textbf{59.6}$_{7.2}$    &  $51.1_{1.9}$    &  \textbf{50.3}$_{3.7}$\\

~ Oracle & {85.5}$_{4.3}$ & {40.5}$_{6.3}$ & {65.2}$_{7.6}$ & {74.7}$_{6.1}$ & {80.4}$_{5.4}$ & {77.3}$_{2.3}$ & {79.4}$_{2.4}$ & {63.3}$_{2.9}$ & {68.4}$_{8.0}$ & {53.9}$_{1.3}$ & {62.5}$_{7.4}$\\

\midrule

GPT-2 0.8B & $74.5_{10.3}$ & $34.7_{8.2}$  &  $55.0_{12.5}$  &  $64.6_{13.1}$   &  $70.9_{12.7}$   &  $65.5_{8.7}$   & $56.4_{9.1}$    & $56.5_{2.7}$    & $62.2_{11.6}$ &  $53.2_{2.0}$   &  $38.8_{8.5}$\\
~ LocalE &  $81.1_{5.5}$  & $40.3_{4.7}$ & $56.7_{7.5}$  &  $82.6_{4.2}$ &    $85.4_{3.8}$ &  $73.6_{4.8}$   &  \textbf{70.4}$_{4.2}$   &  $56.2_{1.7}$  & $62.7_{8.1}$ &  $53.3_{1.6}$   &  $38.4_{5.2}$\\

~ GlobalE    & \textbf{84.8}$_{4.1}$ & \textbf{46.9}$_{1.1}$  &  \textbf{67.7}$_{3.6}$   & \textbf{84.3}$_{2.9}$ &  \textbf{86.7}$_{2.5}$   &   \textbf{75.8}$_{3.1}$ &  $68.6_{6.5}$   &  \textbf{57.2}$_{2.3}$   & \textbf{70.7}$_{3.6}$  &  \textbf{53.5}$_{1.5}$   & \textbf{41.2}$_{4.5}$\\

~ Oracle & {88.9}$_{1.8}$ & {48.4}$_{0.7}$ & {72.3}$_{3.3}$ & {87.5}$_{1.1}$ & {89.9}$_{0.9}$ & {80.3}$_{4.9}$ & {76.6}$_{4.1}$ & {62.1}$_{1.5}$ & {78.1}$_{1.3}$ & {57.3}$_{1.0}$ & {53.2}$_{5.3}$ \\

\midrule

GPT-2 1.5B &  $66.8_{10.8}$ & $41.7_{6.7}$  &  $82.6_{2.5}$   & $59.1_{11.9}$    &  $56.9_{9.0}$   &  $73.9_{8.6}$   & $59.7_{ 10.4}$    & $53.1_{3.3}$    & $77.6_{7.3}$ &  $55.0_{1.4}$   & $53.8_{4.7}$\\
~ LocalE &  $76.7_{8.2}$   & \textbf{45.1}$_{3.1}$ & $83.8_{1.7}$  &  $78.1_{5.6}$  &  $71.8_{8.0}$ & $78.5_{3.6}$    & $69.7_{5.8}$    & $53.6_{3.1}$  & $79.3_{3.7}$ & \textbf{56.8}$_{1.1}$  & $52.6_{3.9}$\\

~ GlobalE    &  \textbf{81.8}$_{3.9}$ & $43.5_{4.5}$  &  \textbf{83.9}$_{1.8}$ & \textbf{77.9}$_{5.7}$ &  \textbf{73.4}$_{6.0}$   &  \textbf{81.4}$_{2.1}$   &  \textbf{70.9}$_{6.0}$   &  \textbf{55.5}$_{3.0}$ &   \textbf{83.9}$_{1.2}$  &  $56.3_{1.2}$   & \textbf{55.1}$_{4.6}$\\

~ Oracle & {86.1}$_{1.5}$ & {50.9}$_{1.0}$ & {87.3}$_{1.5}$ & {84.0}$_{2.7}$ & {80.3}$_{3.3}$ & {85.1}$_{1.4}$ & {79.9}$_{5.7}$ & {59.0}$_{2.3}$ & {86.1}$_{0.7}$ & {58.2}$_{0.6}$ & {63.9}$_{4.3}$\\
\midrule

GPT-3 2.7B & $78.0_{10.7}$    & $35.3_{6.9}$  &  $81.1_{1.8}$ &  $68.0_{12.9}$   & $76.8_{11.7}$    & $66.5_{10.3}$    & $49.1_{2.9}$    & $55.3_{4.4}$    &  $72.9_{4.8}$   &  $48.6_{1.9}$   & $50.4_{0.7}$ \\
~ LocalE & \textbf{81.0}$_{6.0}$    &  $42.3_{4.7}$ & $80.3_{1.7}$   & $75.6_{4.1}$    &   $79.0_{5.5}$  &  $72.5_{5.8}$   & $54.2_{4.2}$ &  $54.0_{2.6}$   &  $72.3_{4.6}$   &  $50.4_{1.9}$   & $50.5_{0.8}$ \\
~ GlobalE &  $80.2_{4.2}$   & \textbf{43.2}$_{4.3}$ & \textbf{81.2}$_{0.9}$  & \textbf{76.1}$_{3.8}$     &    \textbf{80.3}$_{3.4}$ &  \textbf{73.0}$_{4.3}$   & \textbf{54.3}$_{4.0}$ &  \textbf{56.7}$_{2.0}$   &    \textbf{78.1}$_{1.9}$ &  \textbf{51.3}$_{1.8}$   &  \textbf{51.2}$_{0.8}$\\
~ Oracle & {89.8}$_{0.7}$ & {48.0}$_{1.1}$ & {85.4}$_{1.6}$ & {87.4}$_{0.9}$ & {90.1}$_{0.7}$ & {80.9}$_{1.4}$ & {60.3}$_{10.3}$ & {62.8}$_{4.2}$ & {81.3}$_{2.9}$ & {53.4}$_{3.1}$ & {52.5}$_{1.4}$\\

\midrule

GPT-3 175B &   \textbf{93.9}$_{0.6}$  & $54.4_{2.5}$  & $95.4_{0.9}$  &   \textbf{94.6}$_{0.7}$  &  $91.0_{1.0}$   &  $83.2_{1.5}$   &  $71.2_{7.3}$   &  $72.1_{2.7}$   &  $85.1_{1.7}$   &  $70.8_{2.8}$   & $75.1_{5.1}$ \\
~ LocalE  & $93.8_{0.5}$   & \textbf{56.0}$_{1.7}$  & $95.5_{0.9}$ & {94.5}$_{0.7}$     &  {91.3}$_{0.5}$   & \textbf{83.3}$_{1.7}$     &  $75.0_{4.6}$   &   $71.8_{3.2}$  &    \textbf{85.9}$_{0.7}$ &  \textbf{71.9}$_{1.4}$   &  $74.6_{4.2}$\\
~ GlobalE & \textbf{93.9}$_{0.6}$   & $53.2_{2.1}$  & \textbf{95.7}$_{0.7}$  & \textbf{94.6}$_{0.2}$    &   \textbf{91.7}$_{0.4}$  &$82.0_{0.8}$     &   \textbf{76.3}$_{3.5}$  & \textbf{73.6}$_{2.5}$     &   $85.7_{1.0}$   &  $71.8_{1.9}$   &  \textbf{79.9}$_{3.3}$\\

~ Oracle & {94.7}$_{0.2}$ & {58.2} & {96.7}$_{0.2}$ & {95.5}$_{0.2}$ & {92.6}$_{0.4}$ & {85.5}$_{0.8}$ & {81.1}$_{4.9}$ & {77.0}$_{1.2}$ & {87.7}$_{0.6}$ & {74.7}$_{0.4}$ & {83.0}$_{0.9}$\\

\bottomrule
\end{tabular}}
\end{center}

\caption{Our main results on subset of the validation set. To fit the data within the GPT-2 model context window size, we use 1-shot for DBPedia, 2-shot for AGNews, 4-shot for other datasets. All the baseline results are calculated based on 5 different random seeds over 24 train context permutations. LocalE and GlobalE results are calculated based on the top 4 context permutations using our proposed approach. For the GPT-3 175B, we only use 2 seeds with 12 different permutations due to a limited computation budget.}
\label{tab:main_result}
\end{table*}

%% file: tables/change_template.tex
\begin{table}
\begin{center}
\centering
\resizebox{\columnwidth}{!}{%
\begin{tabular}{lcccc}
\toprule
           & Template 1  & Template 2 & Template 3 & Template 4\\
\midrule
GPT-2 0.1B  & 58.9$_{7.8}$ & 57.5$_{6.8}$ & 58.1$_{7.4}$ &  56.6$_{6.6}$\\
~LocalE  & \textbf{65.2}$_{3.9}$ & \textbf{60.7}$_{4.6}$  & \textbf{65.4}$_{4.8}$ & 61.0$_{4.7}$ \\
~GlobalE  & 63.8$_{5.8}$ & 59.0$_{2.9}$ & 64.3$_{4.8}$ &  \textbf{63.5}$_{4.8}$\\
\midrule

GPT-2 0.3B  & 61.0$_{13.2}$ & 63.9$_{11.3}$ & 68.3$_{11.8}$ &  59.2$_{6.4}$\\
~LocalE  & 75.3$_{4.6}$  & 70.0$_{7.2}$ & 80.2$_{4.2}$ & 62.2$_{3.4}$ \\
~GlobalE  & \textbf{78.7}$_{5.2}$ & \textbf{73.3}$_{4.5}$ & \textbf{81.3}$_{4.1}$ &  \textbf{62.8}$_{4.3}$\\
\midrule

GPT-2 0.8B  & 74.5$_{10.3}$ & 66.6$_{10.6}$ & 70.3$_{10.5}$ &  63.7$_{8.9}$\\
~LocalE  & 81.1$_{5.5}$ & 80.0$_{5.6}$ & 73.7$_{6.2}$ & \textbf{71.3}$_{4.5}$ \\
~GlobalE  & \textbf{84.8}$_{4.1}$ & \textbf{80.9}$_{3.6}$ & \textbf{79.8}$_{3.9}$ &  70.7$_{5.3}$\\
\midrule

GPT-2 1.5B  & 66.8$_{10.8}$ & 80.4$_{7.6}$ & 54.5$_{7.9}$ &  69.1$_{10.5}$\\
~LocalE  & 76.7$_{8.2}$ & 83.1$_{3.6}$ & 66.9$_{7.5}$ & 72.7$_{5.5}$ \\
~GlobalE  & \textbf{81.8}$_{3.9}$ & \textbf{83.4}$_{3.2}$ & \textbf{67.2}$_{6.1}$ &  \textbf{74.2}$_{5.3}$\\
\bottomrule
\end{tabular}
}
\end{center}
\caption{Prompt selection performance of different templates on SST-2} 
\label{tab:different-template-result}
\end{table}

%% file: tables/different_template.tex
\begin{table}
\begin{center}
\resizebox{0.9\columnwidth}{!}{%
\begin{tabular}{lll}
\toprule
ID &  Template & Label Mapping\\
\bottomrule
1 & \begin{tabular}[c]{@{}l@{}}Review: \{Sentence\}\\Sentiment: \{Label\}\end{tabular} & positive/negative \\
\midrule
2 & \begin{tabular}[c]{@{}l@{}}Input: \{Sentence\}\\Prediction: \{Label\}\end{tabular} & positive/negative \\
\midrule
3 & \begin{tabular}[c]{@{}l@{}}Review: \{Sentence\}\\Sentiment: \{Label\}\end{tabular} & good/bad \\
\midrule
4 & \begin{tabular}[c]{@{}l@{}} \{Sentence\} It was \{Label\}\end{tabular} & good/bad \\
\bottomrule
\end{tabular}
}
\end{center}
\caption{Different Templates for SST-2}
\label{tab:different_template}
\end{table}

%% file: tables/cross_validation.tex
\begin{table}
\begin{center}
\centering
\resizebox{\columnwidth}{!}{%
\begin{tabular}{lcccc}
\toprule
           & GPT-2 0.1B &  GPT-2 0.3B &  GPT-2 0.8B &  GPT-2 1.5B\\
\midrule
Baseline & $58.9_{7.8}$ & $61.0_{13.2}$ & $74.5_{10.3}$ & $66.8_{10.8}$\\
LocalE & $\textbf{65.2}_{3.9}$ &  $75.3_{4.6}$ & $81.1_{5.5}$ & $76.7_{8.2}$\\
GlobalE & $63.8_{5.8}$ & $\textbf{78.7}_{5.2}$ &  $\textbf{84.8}_{4.1}$ &$\textbf{81.8}_{3.9}$ \\
Split Training Set& $62.8_{5.3}$ & $64.2_{6.1}$  & $75.1_{6.8}$ & $71.4_{7.8}$ \\

\bottomrule
\end{tabular}
}
\end{center}
\caption{Comparing our method with splitting the training set into train and development for SST-2.} 
\label{tab:cross_validation}
\end{table}

%% file: appendix.tex
\input{tables/appendix_prompt_template}

\input{tables/notation}
\input{tables/datasets}

\input{tables/generation-examples}

%% file: tables/appendix_prompt_template.tex
\begin{table*}
\begin{center}
\resizebox{2.0\columnwidth}{!}{%
\begin{tabular}{lll}
\toprule
Dataset &  Prompt & Label Mapping\\
\bottomrule
SST-2 & \begin{tabular}[c]{@{}l@{}}Review: {contains no wit , only labored gags}\\Sentiment: negative\end{tabular} & positive/negative \\
\midrule
SST-5 & \begin{tabular}[c]{@{}l@{}}Review: {apparently reassembled from the cutting-room floor of any given daytime soap .}\\Sentiment: terrible\end{tabular} & terrible/bad/okay/good/great \\
\midrule

MR & \begin{tabular}[c]{@{}l@{}}Review: {lame sweet home leaves no southern stereotype unturned .}\\Sentiment: negative\end{tabular} & negative/positive \\
\midrule

CR & \begin{tabular}[c]{@{}l@{}}Review: {bluetooth does not work on this phone .}\\Sentiment: negative\end{tabular} & negative/positive \\
\midrule
MPQA & \begin{tabular}[c]{@{}l@{}}Review: {dangerous situation}\\Sentiment: negative\end{tabular} & negative/positive \\
\midrule
Subj & \begin{tabular}[c]{@{}l@{}}Input: {too slow , too boring , and occasionally annoying .}\\Type: subjective\end{tabular} & subjective/objective \\
\midrule
TREC & \begin{tabular}[c]{@{}l@{}}Question: {When did the neanderthal man live ?}\\Type: number\end{tabular} & \begin{tabular}[c]{@{}l@{}}description/entity/expression/\\human/location/number\end{tabular}  \\
\midrule
AGNews & \begin{tabular}[c]{@{}l@{}}input: {Wall St. Bears Claw Back Into the Black (Reuters).}\\type: business\end{tabular} & world/sports/business/technology \\
\midrule
DBPedia & \begin{tabular}[c]{@{}l@{}}input: {CMC Aviation is a charter airline based in Nairobi Kenya.}\\type: company\end{tabular} & \begin{tabular}[c]{@{}l@{}}company/school/artist/athlete/politics/\\transportation/building/nature/village/\\animal/plant/album/film/book\end{tabular}  \\
\midrule
CB &  \begin{tabular}[c]{@{}l@{}}premise: It was a complex language. Not written down but handed down. \\\hspace{1.3cm} One might say it was peeled down.\\hypothesis: {the language was peeled down}\\prediction: {true}\end{tabular} &  true/false/neither\\
\midrule
RTE &  \begin{tabular}[c]{@{}l@{}}premise: {No Weapons of Mass Destruction Found in Iraq Yet.}\\hypothesis: {Weapons of Mass Destruction Found in Iraq.}\\prediction: {False}\end{tabular} &  True/False\\
\bottomrule
\end{tabular}
}
\end{center}
\caption{Prompt template and label mapping for different tasks.}
\label{tab:appendix_prompt}
\end{table*}

%% file: tables/notation.tex
\begin{table}
\begin{center}
\centering
\resizebox{\columnwidth}{!}{%
\begin{tabular}{lll}
\toprule
Notation & Description & Examples \\
\midrule
x & sentence & nice movie\\
y & label & positive\\
$\mathcal{T}$(x) & \begin{tabular}[c]{@{}l@{}}template-based transformation\\ without label \end{tabular}  & Review: nice movie \\
$\mathcal{T}$(x,y) & template-based transformation &  \begin{tabular}[c]{@{}l@{}}Review: nice movie\\Sentiment: positive\end{tabular}\\
$\mathcal{T}^{-1}$($\mathcal{T}$(x,y)) &\begin{tabular}[c]{@{}l@{}}extract (sentence, label) pair \\from text sequence \end{tabular} &(nice movie, positive) \\
\bottomrule
\end{tabular}
}
\end{center}
\caption{Examples of transformation notations.}
\label{tab:notation}
\end{table}

%% file: tables/datasets.tex
\begin{table}
\begin{center}
\centering
\resizebox{\columnwidth}{!}{%
\begin{tabular}{lccc}
\toprule
Dataset &  $\#$ of Classes & Avg. Len. & Balanced\\
\midrule
SST-2~\cite{socher2013sst2} & 2 & 12.4 & Yes\\
SST-5~\cite{socher2013sst2} & 5 & 23.1 & No \\
MR~\cite{pang2005mr} & 2 & 25.7 & Yes\\
CR~\cite{hu2004cr} & 2 & 22.1 & Yes\\
MPQA~\cite{wiebe2005mpqa} & 2 & 3.9 & Yes\\
Subj~\cite{pang2004subj} & 2 & 28.9 & Yes\\
TREC~\cite{voorhees2000trec} & 6 & 11.6 & No\\
AGNews~\cite{zhang2015agnewsdbpedia} & 4 &  53.8 & Yes\\
DBPedia~\cite{zhang2015agnewsdbpedia} & 14 & 65.5 & Yes\\
CB~\cite{de2019commitmentbank} &  3 &  69.7/8.4& No\\
RTE~\cite{dagan2005rte} & 2 &  55.3/11.9 & Yes \\
\bottomrule
\end{tabular}
}
\end{center}
\caption{Statistics of evaluation datasets, average length is calculated based on GPT-2 sentence-piece length. For sentence-pair tasks, we report each sentence's average length separately.} 
\label{tab:datasets}
\end{table}

%% file: tables/generation-examples.tex
\begin{table*}
\begin{center}
\resizebox{2\columnwidth}{!}{%
\begin{tabular}{ll}
\toprule
Dataset &  Synthetic data\\
\bottomrule
SST-2 & \begin{tabular}[c]{@{}l@{}}not sure where to even begin\\the only real film on our watch lists\\no one will care because it is just one story\end{tabular} \\
\midrule
SST-5 & \begin{tabular}[c]{@{}l@{}}not a bad documentary, but the story feels tacked on.\\one that i have never liked and was always too long to understand and not enjoyable in parts.\\This movie is the opposite of what it pretentious title implies.\end{tabular} \\
\midrule
DBPedia & \begin{tabular}[c]{@{}l@{}}Gweno Mott's book: Gweno is a New Yorker cartoonist published by Little, Brown, 1995/2002/2013.\\L. Ego Equestrians is North America's first dedicated equine show in Las Vegas.\\Graphed is a graph visualization package from the GraphViz project.\end{tabular} \\
\midrule
MR & \begin{tabular}[c]{@{}l@{}}a solid first film for the debut helmer.\\ A good deal more of the material in his previous films can be found here but this film does not come across [...]\\it is so effective and engaging It feels more real And at some point, maybe it was about [...] \end{tabular} \\ 
\midrule
CR & \begin{tabular}[c]{@{}l@{}}It works just the same, i just prefer my iPhone 6.\\the battery last so long for me it feels like ive already had my phone a year.\\works great with both phones\end{tabular} \\
\midrule
MPQA & \begin{tabular}[c]{@{}l@{}}this is really going nowhere\\why does it look so angry??\\Excellent book and will get a good reputation\end{tabular} \\
\midrule
Subj & \begin{tabular}[c]{@{}l@{}}this will become apparent as it gets older.\\how about something more subtle to show this girl's love?\\a perfect summary of an episode where the entire series is one massive meta romp, with [...]\end{tabular} \\
\midrule
TREC & \begin{tabular}[c]{@{}l@{}}Whales can hold 4 gallons. Whaler can also be written as: What whale is named Whalerel?\\To a certain degree, how do human eyes perceive colour?\\From where does our moon orbit, in Earth's Solar System?\end{tabular} \\
\midrule
AGNews & \begin{tabular}[c]{@{}l@{}}Google buys for \$11bn: A-Z and thesaurus online, music search; photo service and TV site [...]\\Saudi-born billionaire takes \$5 Billion Hit With Bankrupt. Saudi millionaire Sultan Al-Amoudi said [...]\\China's 'Sesame' takes over for South Korea in world TV race as US TV loses market dominance.[...]\end{tabular} \\
\midrule
RTE & \begin{tabular}[c]{@{}l@{}}Premise: The Tuareg are a nomadic people who live in the Sahara desert.\\ Hypothesis: Tuareg are nomadic people who lived in the Sahara desert before the arrival of the Arabs.\\\\ Premise: In the early 1940s, the United States and the Soviet Union were at war with Germany. \\ Hypothesis: Germany was at war with the United States and Russia.\\\\ Premise: Water is a precious commodity.\\ Hypothesis: Water is not a precious commodity.\end{tabular} \\
\midrule
CB & \begin{tabular}[c]{@{}l@{}}Premise: In the back corner of Melissa's classroom her father walked through the door and walked across the front. [...]\\Hypothesis: his curiosity was directed towards some, something other than Melissa\\\\Premise: Maggie took Gloria out for a drive to the nearby city limits of Fort Myers on Tuesday\\Hypothesis: he couldn't bear looking down his nose at all the other houses\\\\Premise: There was one in Dallas. When it came out in New Jersey. And there were,[...]\\Hypothesis: I would never see that movie \end{tabular} \\
\bottomrule
\end{tabular}
}
\end{center}
\caption{Artificial development set generated by GPT2-XL (1.5B). We random select three examples per dataset. Long sentences are trimmed due to limited space.}
\label{tab:generation_examples}
\end{table*}

%% file: acl.bbl
\begin{thebibliography}{28}
\expandafter\ifx\csname natexlab\endcsname\relax\def\natexlab#1{#1}\fi

\bibitem[{Brown et~al.(2020)Brown, Mann, Ryder, Subbiah, Kaplan, Dhariwal,
  Neelakantan, Shyam, Sastry, Askell et~al.}]{brown2020language}
Tom~B Brown, Benjamin Mann, Nick Ryder, Melanie Subbiah, Jared Kaplan, Prafulla
  Dhariwal, Arvind Neelakantan, Pranav Shyam, Girish Sastry, Amanda Askell,
  et~al. 2020.
\newblock Language models are few-shot learners.
\newblock \emph{arXiv preprint arXiv:2005.14165}.

\bibitem[{Dagan et~al.(2005)Dagan, Glickman, and Magnini}]{dagan2005rte}
Ido Dagan, Oren Glickman, and Bernardo Magnini. 2005.
\newblock The pascal recognising textual entailment challenge.
\newblock In \emph{Machine Learning Challenges Workshop}, pages 177--190.
  Springer.

\bibitem[{Davison et~al.(2019)Davison, Feldman, and
  Rush}]{davison2019commonsense}
Joe Davison, Joshua Feldman, and Alexander~M Rush. 2019.
\newblock Commonsense knowledge mining from pretrained models.
\newblock In \emph{Proceedings of the 2019 Conference on Empirical Methods in
  Natural Language Processing and the 9th International Joint Conference on
  Natural Language Processing (EMNLP-IJCNLP)}, pages 1173--1178.

\bibitem[{De~Marneffe et~al.(2019)De~Marneffe, Simons, and
  Tonhauser}]{de2019commitmentbank}
Marie-Catherine De~Marneffe, Mandy Simons, and Judith Tonhauser. 2019.
\newblock The commitmentbank: Investigating projection in naturally occurring
  discourse.
\newblock In \emph{proceedings of Sinn und Bedeutung}, pages 107--124.

\bibitem[{Devlin et~al.(2019)Devlin, Chang, Lee, and
  Toutanova}]{devlin2019bert}
Jacob Devlin, Ming-Wei Chang, Kenton Lee, and Kristina Toutanova. 2019.
\newblock Bert: Pre-training of deep bidirectional transformers for language
  understanding.
\newblock In \emph{Proceedings of the 2019 Conference of the North American
  Chapter of the Association for Computational Linguistics: Human Language
  Technologies, Volume 1 (Long and Short Papers)}, pages 4171--4186.

\bibitem[{Gao et~al.(2020)Gao, Fisch, and Chen}]{gao2020making}
Tianyu Gao, Adam Fisch, and Danqi Chen. 2020.
\newblock Making pre-trained language models better few-shot learners.
\newblock \emph{arXiv preprint arXiv:2012.15723}.

\bibitem[{Hu and Liu(2004)}]{hu2004cr}
Minqing Hu and Bing Liu. 2004.
\newblock Mining and summarizing customer reviews.
\newblock In \emph{Proceedings of the tenth ACM SIGKDD international conference
  on Knowledge discovery and data mining}, pages 168--177.

\bibitem[{Jiang et~al.(2020)Jiang, Xu, Araki, and Neubig}]{jiang2020can}
Zhengbao Jiang, Frank~F Xu, Jun Araki, and Graham Neubig. 2020.
\newblock How can we know what language models know?
\newblock \emph{Transactions of the Association for Computational Linguistics},
  8:423--438.

\bibitem[{Kumar et~al.(2016)Kumar, Irsoy, Ondruska, Iyyer, Bradbury, Gulrajani,
  Zhong, Paulus, and Socher}]{kumar2016ask}
Ankit Kumar, Ozan Irsoy, Peter Ondruska, Mohit Iyyer, James Bradbury, Ishaan
  Gulrajani, Victor Zhong, Romain Paulus, and Richard Socher. 2016.
\newblock Ask me anything: Dynamic memory networks for natural language
  processing.
\newblock In \emph{International conference on machine learning}, pages
  1378--1387. PMLR.

\bibitem[{Liu et~al.(2021)Liu, Shen, Zhang, Dolan, Carin, and
  Chen}]{liu2021makes}
Jiachang Liu, Dinghan Shen, Yizhe Zhang, Bill Dolan, Lawrence Carin, and Weizhu
  Chen. 2021.
\newblock What makes good in-context examples for gpt-$3 $?
\newblock \emph{arXiv preprint arXiv:2101.06804}.

\bibitem[{Liu et~al.(2019)Liu, Ott, Goyal, Du, Joshi, Chen, Levy, Lewis,
  Zettlemoyer, and Stoyanov}]{liu2019roberta}
Yinhan Liu, Myle Ott, Naman Goyal, Jingfei Du, Mandar Joshi, Danqi Chen, Omer
  Levy, Mike Lewis, Luke Zettlemoyer, and Veselin Stoyanov. 2019.
\newblock Roberta: A robustly optimized bert pretraining approach.
\newblock \emph{arXiv preprint arXiv:1907.11692}.

\bibitem[{Pang and Lee(2004)}]{pang2004subj}
Bo~Pang and Lillian Lee. 2004.
\newblock A sentimental education: Sentiment analysis using subjectivity
  summarization based on minimum cuts.
\newblock In \emph{Proceedings of the 42nd Annual Meeting of the Association
  for Computational Linguistics (ACL-04)}, pages 271--278.

\bibitem[{Pang and Lee(2005)}]{pang2005mr}
Bo~Pang and Lillian Lee. 2005.
\newblock Seeing stars: Exploiting class relationships for sentiment
  categorization with respect to rating scales.
\newblock In \emph{Proceedings of the 43rd Annual Meeting of the Association
  for Computational Linguistics (ACL’05)}, pages 115--124.

\bibitem[{Paulus et~al.(2018)Paulus, Xiong, and Socher}]{paulus2018deep}
Romain Paulus, Caiming Xiong, and Richard Socher. 2018.
\newblock A deep reinforced model for abstractive summarization.
\newblock In \emph{International Conference on Learning Representations}.

\bibitem[{Perez et~al.(2021)Perez, Kiela, and Cho}]{perez2021true}
Ethan Perez, Douwe Kiela, and Kyunghyun Cho. 2021.
\newblock True few-shot learning with language models.
\newblock \emph{arXiv preprint arXiv:2105.11447}.

\bibitem[{Peters et~al.(2018)Peters, Neumann, Iyyer, Gardner, Clark, Lee, and
  Zettlemoyer}]{peters2018deep}
Matthew Peters, Mark Neumann, Mohit Iyyer, Matt Gardner, Christopher Clark,
  Kenton Lee, and Luke Zettlemoyer. 2018.
\newblock Deep contextualized word representations.
\newblock In \emph{Proceedings of the 2018 Conference of the North American
  Chapter of the Association for Computational Linguistics: Human Language
  Technologies, Volume 1 (Long Papers)}, pages 2227--2237.

\bibitem[{Petroni et~al.(2020)Petroni, Lewis, Piktus, Rockt{\"a}schel, Wu,
  Miller, and Riedel}]{petroni2020context}
Fabio Petroni, Patrick Lewis, Aleksandra Piktus, Tim Rockt{\"a}schel, Yuxiang
  Wu, Alexander~H Miller, and Sebastian Riedel. 2020.
\newblock How context affects language models' factual predictions.
\newblock In \emph{Automated Knowledge Base Construction}.

\bibitem[{Petroni et~al.(2019)Petroni, Rockt{\"a}schel, Riedel, Lewis, Bakhtin,
  Wu, and Miller}]{petroni-etal-2019-language}
Fabio Petroni, Tim Rockt{\"a}schel, Sebastian Riedel, Patrick Lewis, Anton
  Bakhtin, Yuxiang Wu, and Alexander Miller. 2019.
\newblock \href {https://doi.org/10.18653/v1/D19-1250} {Language models as
  knowledge bases?}
\newblock In \emph{Proceedings of the 2019 Conference on Empirical Methods in
  Natural Language Processing and the 9th International Joint Conference on
  Natural Language Processing (EMNLP-IJCNLP)}, pages 2463--2473, Hong Kong,
  China. Association for Computational Linguistics.

\bibitem[{Radford et~al.(2019)Radford, Wu, Child, Luan, Amodei, and
  Sutskever}]{Radford2019LanguageMA}
A.~Radford, Jeffrey Wu, R.~Child, David Luan, Dario Amodei, and Ilya Sutskever.
  2019.
\newblock Language models are unsupervised multitask learners.
\newblock In \emph{OpenAI Blog}.

\bibitem[{Raffel et~al.(2020)Raffel, Shazeer, Roberts, Lee, Narang, Matena,
  Zhou, Li, and Liu}]{raffel2020exploring}
Colin Raffel, Noam Shazeer, Adam Roberts, Katherine Lee, Sharan Narang, Michael
  Matena, Yanqi Zhou, Wei Li, and Peter~J Liu. 2020.
\newblock Exploring the limits of transfer learning with a unified text-to-text
  transformer.
\newblock \emph{Journal of Machine Learning Research}, 21:1--67.

\bibitem[{Schick and Sch{\"u}tze(2020)}]{schick2020pet}
Timo Schick and Hinrich Sch{\"u}tze. 2020.
\newblock It's not just size that matters: Small language models are also
  few-shot learners.
\newblock \emph{arXiv preprint arXiv:2009.07118}.

\bibitem[{Shin et~al.(2020)Shin, Razeghi, Logan~IV, Wallace, and
  Singh}]{shin2020autoprompt}
Taylor Shin, Yasaman Razeghi, Robert~L Logan~IV, Eric Wallace, and Sameer
  Singh. 2020.
\newblock Autoprompt: Eliciting knowledge from language models with
  automatically generated prompts.
\newblock \emph{arXiv preprint arXiv:2010.15980}.

\bibitem[{Socher et~al.(2013)Socher, Perelygin, Wu, Chuang, Manning, Ng, and
  Potts}]{socher2013sst2}
Richard Socher, Alex Perelygin, Jean Wu, Jason Chuang, Christopher~D Manning,
  Andrew~Y Ng, and Christopher Potts. 2013.
\newblock Recursive deep models for semantic compositionality over a sentiment
  treebank.
\newblock In \emph{Proceedings of the 2013 conference on empirical methods in
  natural language processing}, pages 1631--1642.

\bibitem[{Voorhees and Tice(2000)}]{voorhees2000trec}
Ellen~M Voorhees and Dawn~M Tice. 2000.
\newblock Building a question answering test collection.
\newblock In \emph{Proceedings of the 23rd annual international ACM SIGIR
  conference on Research and development in information retrieval}, pages
  200--207.

\bibitem[{Wiebe et~al.(2005)Wiebe, Wilson, and Cardie}]{wiebe2005mpqa}
Janyce Wiebe, Theresa Wilson, and Claire Cardie. 2005.
\newblock Annotating expressions of opinions and emotions in language.
\newblock \emph{Language resources and evaluation}, 39(2):165--210.

\bibitem[{Yang et~al.(2019)Yang, Dai, Yang, Carbonell, Salakhutdinov, and
  Le}]{yang2019xlnet}
Zhilin Yang, Zihang Dai, Yiming Yang, Jaime Carbonell, Ruslan Salakhutdinov,
  and Quoc~V Le. 2019.
\newblock Xlnet: Generalized autoregressive pretraining for language
  understanding.
\newblock \emph{arXiv preprint arXiv:1906.08237}.

\bibitem[{Zhang et~al.(2015)Zhang, Zhao, and Lecun}]{zhang2015agnewsdbpedia}
Xiang Zhang, Junbo Zhao, and Yann Lecun. 2015.
\newblock Character-level convolutional networks for text classification.
\newblock \emph{Advances in Neural Information Processing Systems},
  2015:649--657.

\bibitem[{Zhao et~al.(2021)Zhao, Wallace, Feng, Klein, and
  Singh}]{zhao2021calibrate}
Tony~Z Zhao, Eric Wallace, Shi Feng, Dan Klein, and Sameer Singh. 2021.
\newblock Calibrate before use: Improving few-shot performance of language
  models.
\newblock \emph{arXiv preprint arXiv:2102.09690}.

\end{thebibliography}
